\documentclass{article}
\usepackage{spconf,amsmath,graphicx}

\def\0{{\mathbf 0}}
\def\1{{\mathbf 1}}

\def\f{{\mathbf f}}

\def\l{{\mathbf l}}

\def\n{{\mathbf n}}

\def\v{{\mathbf v}}

\def\x{{\mathbf x}}
\def\y{{\mathbf y}}

\def\D{{\mathbf D}}

\def\F{{\mathbf F}}

\def\F{{\mathbf F}}
\def\H{{\mathbf H}}

\def\L{{\mathbf L}}

\def\U{{\mathbf U}}
\def\V{{\mathbf V}}
\def\W{{\mathbf W}}

\def\ie{{\textit{i.e.}}}

\def\cE{{\mathcal E}}

\def\cG{{\mathcal G}}

\def\cV{{\mathcal V}}

\def\balpha{{\boldsymbol \alpha}}

\def\bGamma{{\boldsymbol \Gamma}}
\def\bLambda{{\boldsymbol \Lambda}}

\usepackage{mathtools}
\usepackage{amsmath,amsopn,amssymb}
\usepackage{amsthm} 
\usepackage{graphicx,xspace,color,soul}
\usepackage{epsfig}
\usepackage{longtable,multirow}
\usepackage{array,float}
\usepackage{algorithmic}
\usepackage[linesnumbered, ruled]{algorithm2e}
\usepackage{bm,epstopdf}
\usepackage{tabularx}
\usepackage{afterpage}
\usepackage{booktabs}
\usepackage{multirow}
\usepackage{caption}
\usepackage{subcaption}
\usepackage[export]{adjustbox}

\usepackage{enumitem}

\newcolumntype{Y}{>{\centering\arraybackslash}X}

\renewcommand{\arraystretch}{1.3}
\newcommand{\ra}[1]{\renewcommand{\arraystretch}{#1}}

\def\balpha{{\boldsymbol \alpha}}

\def\bLambda{{\boldsymbol \Lambda}}

\DeclareMathOperator*{\argmin}{arg\,min}
\makeatletter
\DeclareRobustCommand\onedot{\futurelet\@let@token\@onedot}
\def\@onedot{\ifx\@let@token.\else.\null\fi\xspace}
 
\def\ie{\emph{i.e}\onedot}

\makeatother

\title{Unrolling of Deep Graph Total Variation for Image Denoising}
\name{Huy Vu$^\dagger$, Gene Cheung$^\dagger$, Yonina C. Eldar$^\ddagger$
\thanks{Gene Cheung acknowledges the support of the NSERC grants RGPIN-2019-06271,  RGPAS-2019-00110.}
}
\address{$^\dagger$Department of Electrical Engineering \& Computer Science, York University, Toronto, Canada\\
         $^\ddagger$Faculty of Mathematics and Computer Science, Weizmann Institute of Science, Rehovot, Israel}

\begin{document}
\ninept
\maketitle
\begin{abstract}
While deep learning (DL) architectures like convolutional neural networks (CNNs) have enabled effective solutions in image denoising, in general their implementations overly rely on training data, lack interpretability, and require tuning of a large parameter set. 
In this paper, we combine classical graph signal filtering with deep feature learning into a competitive hybrid design---one that utilizes interpretable analytical low-pass graph filters and employs 80\% fewer network parameters than state-of-the-art DL denoising scheme DnCNN.
Specifically, to construct a suitable similarity graph for graph spectral filtering, we first adopt a CNN to learn feature representations per pixel, and then compute feature distances to establish edge weights.
Given a constructed graph, we next formulate a convex optimization problem for denoising using a graph total variation (GTV) prior.
Via a $l_1$ graph Laplacian reformulation, we interpret its solution in an iterative procedure as a graph low-pass filter and derive its frequency response. 
For fast filter implementation, we realize this response using a Lanczos approximation. Experimental results show that in the case of statistical mistmatch, our algorithm outperformed DnCNN by up to 3dB in PSNR.

\end{abstract}
\begin{keywords}
image denoising, graph signal processing, deep learning
\end{keywords}

\vspace{-0.1in}
\section{Introduction}
\label{sec:intro}
Image denoising is a basic image restoration problem, where the goal is to recover the original image (or image patch) $\x$ given only a noisy observation $\y$. 
Modern image denoising methods can be roughly categorized into model-based and learning-based. 
Model-based methods \cite{rudin92,elad06, dabov07,beck09denoising} rely on signal priors like \textit{total variation} (TV) \cite{rudin92, beck09denoising} and sparse representation \cite{elad06} to regularize an inherently ill-posed problem.
However, the assumed priors are simplistic and thus do not lead to satisfactory performance in complicated scenarios. 
In contrast, learning-based methods leverage powerful learning abilities of recent deep learning (DL) architectures such as \textit{convolutional neural networks} (CNN) to compute mappings directly from $\y$ to $\x$, given a large training dataset \cite{zhang17dncnn, Vemulapalli2016DeepGC, Tai2017MemNetAP}. 
These methods are overly dependent on the training data (with risk of \textit{overfitting}), often resulting in unexplainable filters that require tuning of a large set of network parameters.
Large network parameter size is a significant impediment to practical implementation on platforms like mobile phones that require small memory footprints.

To reduce network parameters, we turn to \textit{graph signal processing} (GSP) \cite{cheung2018graph, ortega18}---the study of signals that reside on combinatorial graphs.
Progress in GSP has led to a family of graph filtering tools tailored for different imaging applications, including compression \cite{hu15,  su17}, denoising \cite{pang17, zeng19, su20}, dequantization of JPEG images 
\cite{liu17lerag} 
and deblurring \cite{bai18}. 
Like early spatial filter work for texture recognition \cite{hadji17}, 
analytically derived filters mean filter coefficients do not require learning, thus reducing the number of network parameters.

Such graph-based techniques provide explainable \textit{spectral interpretation} of their filters, complete with a defined frequency response $f(\lambda)$ in the graph frequency domain. 
For example, a low-pass (LP) graph filter can be derived from \textit{graph Laplacian regularizer} (GLR) \cite{pang17} that assumes a signal $\x$ is smooth with respect to (wrt) a graph, minimizing a regularization term $\x^{\top} \L \x$, where $\L$ is the graph Laplacian matrix. 
\label{eq:GLR_filter}
Other graph signal priors such as \textit{graph total variation} (GTV) \cite{elmoataz08, couprie13, berger17} are also possible.

Since a digital image is a signal on a 2D grid, the key to good graph filtering performance for imaging is the selection of an appropriate underlying graph. 
The conventional choice for edge weight $w_{ij}$ between pixels (nodes) $i$ and $j$ in graph spectral image processing \cite{pang17,liu17lerag,bai18} is based on \textit{bilateral filtering} \cite{bilateral}: an exponential kernel $\exp(\cdot)$ of the negative inter-pixel distance $\|\l_i - \l_j\|^2_2$ and pixel intensity difference $(x_i - x_j)^2$, resulting in a non-negative edge weight $0 \leq w_{ij} \leq 1$. 
This choice of edge weights means that graph connectivity is dependent on the signal, resulting in a \textit{signal-dependent} GLR (SDGLR), $\x^{\top} \L(\x) \x$, which promotes \textit{piecewise smoothness} (PWS) in the reconstructed signal \cite{pang17,liu17lerag}. 
However, this choice is handcrafted and sub-optimal in general.

To learn a good similarity graph, \textit{Deep GLR} (DGLR) \cite{zeng19} retained GLR's analytical filter response $f(\lambda)$ but learned suitable feature vector $\f_i$ per pixel $i$ via CNN
in an end-to-end manner, so that graph edge weight $w_{ij}$ can be computed using feature distance $\|\f_i - \f_j\|^2_2$. 
Because feature vector $\f_i$ is of low dimension, network parameters required are relatively few. 
This hybrid of graph spectral filtering and DL architecture for feature representation learning has demonstrated competitive denoising performance with state-of-the-art methods such as DnCNN \cite{zhang17dncnn}. 
A recent variant called \textit{deep analytical graph filter} (DAGF) \cite{su20} replaced the derived LP filter with \textit{GraphBio} \cite{narang13graphbio}---a biorthogonal graph wavelet with perfect reconstruction property. 
DAGF is faster than DGLR with slightly worse denoising performance. 

    
Extending these recent works on hybrid graph spectral filtering / DL architectures \cite{zeng19,su20}, in this paper we propose a fast image denoising algorithm called \textit{Deep GTV} (DGTV) via unrolling of signal-dependent GTV (SDGTV).
Similar to \cite{zeng19,su20}, we also learn appropriate per pixel feature vectors $\f_i$ to first construct a graph with few network parameters. 
We formulate the denoising problem using SDGTV for regularization, which promotes PWS in reconstructed signals faster than SDGLR \cite{bai18}. 
However, GTV is harder to optimize due to the non-differentiable $l_1$-norm. 
In response, we solve the optimization iteratively, where in each iteration, we rewrite GTV into a quadratic term using a $l_1$ graph Laplacian matrix \cite{bai18}, resulting in an analytical graph spectral filter.
This interpretation allows us to approximate the filter response with a fast Lanczos approximation \cite{susnjara15}, implemented as a layer in a neural network, outperforming the popular Chebychev approximation \cite{onuki17, shuman18}.

Recent works in algorithm unrolling \cite{monga19unroll} include also graph-signal denoising \cite{chen20denoising}.
However, the goal in \cite{chen20denoising} is to learn the most suitable prior for signal denoising on a \textit{fixed} pre-determined graph, while we focus on image denoising, where learning a good graph is a key challenge.

Compared to classical model-based methods like BM3D \cite{dabov07}, our algorithm has better denoising performance thanks to DL architecture's power in feature representation learning. 
Compared to pure DL-based DnCNN \cite{zhang17dncnn}, we reduce network parameter size drastically by 80\% due to our deployment of analytical graph filters that are also interpretable\footnote{A contemporary DL work for graph-based image denoising \cite{valsesia20dgcn} also require tuning of a large network parameter set.}. 
Compared to DGLR \cite{zeng19}, our algorithm reconstructs more PWS images for the same network complexity.
Moreover, in case of statistical mismatch between training and test data, we show that DGTV outperformed DnCNN by up to $3$dB in PSNR. 




\section{Preliminaries}
\label{sec:pre}
An 8-neighborhood graph $\mathcal G = (\mathcal V, \mathcal E, \W	)$ with nodes $\cV$ and edges $\cE$ is constructed to represent an $N$-pixel image patch. 
Each pixel $i$ of the image is represented by a node $i \in \cV$. 
Denote by $w_{ij}$ the weight of an edge $(i,j) \in \cE$ connecting nodes $i$ and $j$. 
Edge weights in $\mathcal G$ are defined in an \textit{adjacency matrix} $\W$, where $\W_{ij} = w_{ij}$ for $(i, j) \in \mathcal E$. 
These edge weights represent pairwise similarities between pixels, \ie, a larger weight $w_{ij}$ indicates that samples at nodes $i$ and $j$ are more similar. 

Given $\W$, a diagonal \textit{degree matrix} $\D$ is defined as $\D_{ii} = \sum_j w_{ij}$. 
A combinatorial \textit{graph Laplacian matrix} $\L$ \cite{ortega18} is defined as $\L = \D - \W$. 
Assuming $w_{ij} \geq 0, \forall (i,j) \in \cE$, $\L$ is provably a \textit{positive semi-definite} (PSD) matrix \cite{cheung2018graph}. 
Given $\L$ is real and symmetric, one can eigen-decompose $\L$ into
$\L = \V \bLambda \V^{\top}$, where columns of $\V$ are the eigenvectors, and $\bLambda = \text{diag}(\lambda_1, \ldots, \lambda_N)$ is a diagonal matrix with eigenvalues (graph frequencies) $\lambda_k$ along its diagonals.
$\V^{\top}$ is called the \textit{graph Fourier transform} (GFT)  \cite{ortega18} that transforms a graph signal $\x$ from the nodal domain to the graph frequency domain via $\balpha = \V^{\top} \x$.

A popular graph signal prior for signal $\x$ is \textit{graph Laplacian regularizer} \cite{pang17} (GLR), defined as 
\begin{align}
\x^\top \L \x = \sum_{(i,j) \in \cE} w_{ij}(x_j - x_i)^2 
= \sum_{k=1}^N \lambda_k \alpha_k^2
\label{eq:glr}
\end{align} 
where $\lambda_k$ are the eigenvalues of $\L$ and $\alpha_k$ are the GFT coefficients of signal $\x$.
Minimizing \eqref{eq:glr} means reducing the signal $\x$'s energy in the high graph frequencies, \ie, low-pass filtering.
Since $\L$ is PSD, GLR is lower-bounded by $0$, $\forall \x \in \mathbb{R}^N$. 
Another popular signal prior is \textit{Graph Total Variation} (GTV) \cite{bai18, couprie13}, defined as 
\begin{align}\| \x \|_{\text{GTV}} = \sum_{(i,j) \in \cE} w_{ij}|x_j - x_i|.\end{align}
GTV is also lower-bounded by $0$ when $w_{ij} \geq 0, \forall (i,j) \in \cE$. 

\section{Algorithm Development}
\label{sec:algo}
\subsection{Architecture Overview}

\begin{figure}[ht]
\centering
\includegraphics[width=1\linewidth]{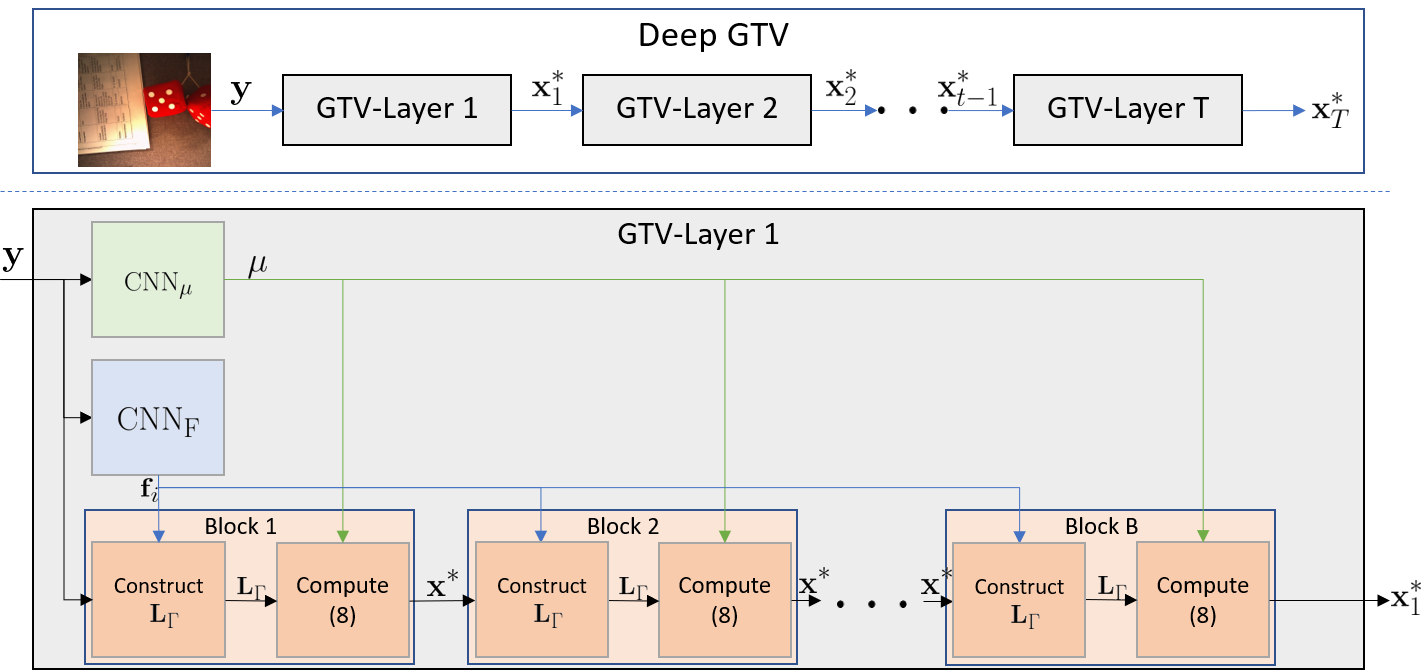}
\caption{Overview of the proposed architecture. Top: Deep GTV composed of multiple GTV-Layer; Bottom: architecture of a GTV-Layer.}
\label{fig:network}
\end{figure}

Our network architecture, as shown in Fig.\;\ref{fig:network}, is composed of $T$ \textit{layers}, each implementing an algorithm iteration.
Each layer is implemented as a cascade of $B$ \textit{blocks}.
At runtime, each block takes as input a noisy input image patch $\y$ and computes a denoised patch $\x$ of the same dimension. 
To assist in the denoising process, in each layer two CNNs are used to compute feature vector $\f_i$ per pixel and a weight parameter $\mu$.
Parameters of the CNNs are trained end-to-end offline using training data.
We next describe the denoising algorithm we unroll into layers and the two CNNs in order.


\subsection{Denoising with Graph Total Variation}	
	
Consider a general image formation model for an additive noise  corrupting an image patch: 
\begin{align}
\y = \x + \n,
\end{align}
where $\y \in \mathbb{R}^N$ is the corrupted observation, $\x \in \mathbb{R}^N$ is the original image patch, and $\n \in \mathbb{R}^N$ is a noise term. 
Our goal is to estimate $\x$ given only $\y$. 
Given a graph $\cG$ with edge weights $w_{ij}$ (to be discussed), we formulate an optimization problem using GTV as a signal prior \cite{bai18} to reconstruct $\x$ given noisy observation $\y$ as
\begin{align}
\min_{\x} \|\y - \x \|^2_2 + 
\mu \sum_{(i,j) \in \cE} 
w_{i,j} |x_i - x_j |,
\label{eq:gtv}
\end{align}
where $\mu > 0$ is a parameter trading off the fidelity and prior terms. 
Problem \eqref{eq:gtv} is convex and can be solved using iterative methods such as \textit{proximal gradient} (PG) \cite{couprie13} but does not have a closed-form solution. 
Moreover, there is no spectral interpretation of the obtained solution.

Instead, as done in \cite{bai18} we transform the GTV $l_1$-norm term in \eqref{eq:gtv} into a quadratic term as follows.
We first define a new adjacency matrix ${\bGamma}$ with edge weights $\Gamma_{ij}$, given a signal estimate $\x^o$, as
\begin{align}
\Gamma_{ij} = \frac{w_{i,j}}{\max \{ |x_i^o - x_j^o | , \rho \}},
\label{eq:Gamma}
\end{align}
where $\rho > 0$ is a chosen parameter to circumvent numerical instability when $|x_i^o - x_j^o| \approx 0$.
Assuming $|x_i^o - x_j^o|$ is sufficiently large and estimate $\x^o$ sufficiently close to signal $\x$, one can see that 
\begin{align}
\Gamma_{i,j} \left( x_i - x_j \right)^2 &=
\frac{w_{ij}}{|x^o_i - x^o_j|} \left(x_i - x_j\right)^2 \approx
w_{ij} \left| x_i - x_j \right|.
\nonumber 
\end{align}
This means that using $\bGamma$, we can express GTV in quadratic form.
Specifically, we define an $l_1$-\textit{Laplacian matrix} $\L_{\Gamma}$ as 
\begin{align}
\mathbf{L}_{\Gamma} = \text{diag} (\bGamma \mathbf{1}) - \bGamma, \label{eq:l1}
\end{align}
where $\1 \in \mathbb{R}^N$ is a length-$N$ vector of all one's.
We then reformulate \eqref{eq:gtv} using a quadratic regularization term given $\L_{\Gamma}$:
\begin{align}
\x^* = \argmin_{\x} & \|\y - \x \|^2_2 + \mu \ \x^\top \mathbf{L}_{\Gamma} \x.
\label{eq:gtv2}
\end{align}

For given estimate $\x^o$ and thus Laplacian $\L_{\Gamma}$, \eqref{eq:gtv2} has a closed-form solution
\begin{align}
\x^* = (\mathbf{I} + \mu \ \mathbf{L}_{\Gamma})^{-1} \y.
\label{eq:sol}
\end{align}
To solve \eqref{eq:gtv}, \eqref{eq:sol} must be computed multiple times, where in each iteration $b+1$, the solution $\x^*_{b} $ from the previous iteration $b$ is used to update edge weights \eqref{eq:Gamma} in $\bGamma$.
In our architecture, each block solves \eqref{eq:sol} once, and a layer composing of $B$ blocks solves \eqref{eq:gtv}. Cascading $T$ layers forms our DGTV architecture, shown in Fig.\;\ref{fig:network}.


\subsection{Feature and Weight Parameter Learning with CNNs}
    
In \eqref{eq:gtv}, a graph $\cG$ with edge weights $w_{ij}$ is assumed.
As done in \cite{zeng19,su20}, in each layer we use a $\text{CNN}_{\F}$ to compute an appropriate feature vector $\f_i \in \mathbb{R}^K$ at runtime for each pixel $i$ in an $N$-pixel patch, using which we construct a graph $\cG$.
Specifically, given feature vectors $\f_i$ and $\f_j$ of pixels (nodes) $i$ and $j$, we compute a non-negative edge weight $w_{ij}$ between them using a Gaussian kernel, \ie, 
\begin{small}
\begin{align}
w_{ij} =
\exp\left( - \frac{\sum_{k=1}^K \left ( \f_i^k - \f_j^k \right )^2 }{\epsilon^2}\right), \label{eq:weight}
\end{align}
\end{small}\noindent
where $\epsilon > 0$ is a parameter. 
To learn parameters in $\text{CNN}_{\F}$ end-to-end that computes $\f_i$, since $\f_i$ is used to construct $\L_{\Gamma}$ and $\L_{\Gamma}$ appears in \eqref{eq:sol}, we can compute the partial derivative of the \textit{mean square error} (MSE) between
the recovered patch $\x^*$ and the ground-truth patch $\x$ wrt $\L_{\Gamma}$ and back-propagate it to update parameters in $\text{CNN}_{\F}$. 
MSE is defined as
\begin{align}
\mathcal L_{\text{MSE}} = \frac{1}{N} \sum_{i=1}^N (x^*_{i} -  x_i)^2.
\end{align} 
Computed edge weights $w_{ij}$ compose the adjacency matrix $\W$. 	


In each layer, we use another CNN---denoted by $\text{CNN}_{\mu}$---to compute weight parameter $\mu$ for a given noisy patch $\y$. 
Similar to $\text{CNN}_{\F}$, $\mu$ appears in \eqref{eq:sol}, and thus we can learn $\text{CNN}_{\mu}$ end-to-end via back-propagation. 

    

{Each GTV-Layer in DGTV filters an image patch by solving \eqref{eq:gtv}, and hence we can learn all CNNs in DGTV in an end-to-end manner by supervising only the error between the final restored patch $\x_T^*$ and the ground-truth patch.}

\subsection{Fast Filter Implementation via Lanczos Approximation}

Computing \eqref{eq:sol} requires matrix inversion.
We can achieve fast execution from the frequency filtering perspective.
By eigen-decomposing $\L_{\Gamma} = \U \bLambda \U^{\top}$, we can interpret 
\eqref{eq:sol} as a LP graph spectral filter with frequency response $f(\lambda)$:
\begin{align}
\x^* &= \U f(\bLambda) \U^{\top} \y \label{eq:gtvFilterOrg} \\
f(\bLambda) &= \text{diag}\left( (1 + \mu \lambda_1))^{-1}, \ldots, (1 + \mu \lambda_N)^{-1} \right).
\label{eq:gtvFilter}
\end{align}
The function in \eqref{eq:gtvFilter} is LP, because for large graph frequency $\lambda$, the frequency response $f(\lambda) = 1/(1 + \mu \lambda)$ is smaller.
Given $f(\bLambda)$, we can avoid matrix inversion by using accelerated graph filter implementations. 
Chebyshev polynomials approximation \cite{onuki17, shuman18} and Lanczos method \cite{susnjara15} are existing filter approximation methods in the literature. 


For graph filter acceleration, we chose the Lanczos method \cite{susnjara15} for our implementation.
First, we compute an orthonormal basis $\V_M = [\v_1,\dots,\v_M]$ of the Krylov subspace $K_M(\L_{\Gamma}, \y) = \text{span}(\y, \L_{\Gamma}\y,\dots,\L_{\Gamma}^{M-1}\y)$ using the Lanczos method. 
The following symmetric tridiagonal matrix $\H_M$ relates $\V_M$ and $\L_{\Gamma}$, where $\alpha_m$ and $\beta_m$ are scalars computed by the Lanczos algorithm.

\vspace{-0.1in}
\begin{footnotesize}
\begin{align}
\H_M = \V_M^* \L_{\Gamma} \V_M = \begin{bmatrix}
\alpha_1 & \beta_2 &  &  & \\ 
\beta_2 & \alpha_2 & \beta_3  &  &\\ 
 & \beta_3 & \alpha_3  &  \ddots&\\
  &  & \ddots& \ddots  &  \beta_M\\
   &  &  & \beta_M & \alpha_M
\end{bmatrix}.
    \end{align}
\end{footnotesize}

The computational cost of the algorithm is $\mathcal O (M  | \mathcal E |)$. 
The following approximation to the graph filter in \eqref{eq:sol} was proposed in \cite{gallopoulos92}:
\begin{align}
f(\L)\y \approx  \| \y \|_2 V_M f(\H_M) e_1 := f_M,
\label{eq:lanc}
\end{align}
where $e_1$ is the first unit vector. 
Due to the eigenvalue interlacing property, the eigenvalues of $\H_M$ are inside the interval $[\lambda_1, \lambda_N] $, and thus $f(\H_M)$ is well-defined. 
The evaluation of \eqref{eq:lanc} is inexpensive since $M \ll N $, leading to accelerated implementation of $f(\L)\y$.

\section{Experiments}
\label{sec:results}
    
\subsection{Comparison between approximation methods}

\begin{figure}[H]
\begin{subfigure}[h]{0.24\textwidth}
\centering
\includegraphics[width=\textwidth]{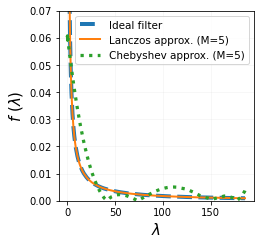}
\end{subfigure}
\hfill
\begin{subfigure}[h]{0.238\textwidth}
\centering
\includegraphics[width=\textwidth]{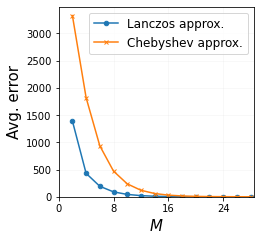}
\end{subfigure}
\vspace{-0.1in}
\caption{Left: sample frequency response of Lanczos and Chebyshev method at $M=5$; Right: average approximation errors wrt $M$. }
\label{fig:exp02}
\end{figure}

We first conducted experiments to compare the Lanczos method with the Chebyshev method in approximating the graph filter frequency response in \eqref{eq:gtvFilter}. 
Specifically, we ran the graph filter \eqref{eq:gtvFilter} on $1000$ random inputs using three filter implementations: i) the original filter \eqref{eq:gtvFilter}, ii) Lanczos approximation with order $M$, and ii) Chebyshev approximation with order $M$. 
We measured the average MSE between the outputs of the original filter and the two approximation methods. 
The results are shown in Fig.\;\ref{fig:exp02}: the left plot shows a sample frequency response of the LP filter for $M=5$, and the right plot shows the approximation errors as a function of $M$. 
We observe that the Lanczos approximation outperformed the Chebychev approximation significantly, especially for small $M$.
We note again that these approximations are possible thanks to the analytical frequency response we derived in \eqref{eq:gtvFilter}.

    
\subsection{Denoising Performance Comparison}
	
\begin{figure}[t]
\centering
\includegraphics[width=.98\linewidth]{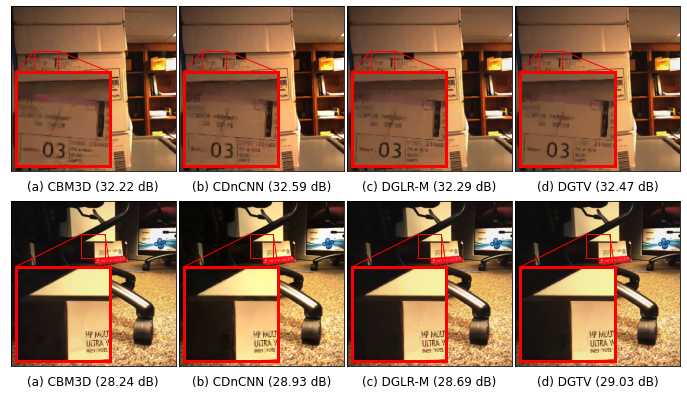}
\vspace{-0.1in}
\caption{Sample results of AWGN denoising. Training and testing on $\sigma=25$. }
\label{fig:gauss}
\end{figure}
    

We compared our proposed DGTV against several state-of-the-art image denoising schemes: a well-known model-based method BM3D \cite{dabov07}, a pure deep learning method CDnCNN \cite{zhang17dncnn}, and DGLR \cite{zeng19}---a hybrid of CNN and analytical graph filters derived from a convex optimization formulation regularized using GLR.
	
\textit{Dataset}. 
We used a small subset of the RENOIR Dataset \cite{renoir} for experiments.
The subset selection criterion is described in Section 4.2 of \cite{zeng19}. 
This subset contained 10 pairs of high resolution images captured by a smartphone. 
Each pair contained a clean image and a noisy image (caused by low-light capturing). 
We used only the clean images, on which we added artificial Gaussian noise in our experiments. 
All images were $720 \times 720$ pixels and each was divided into patches of size $36 \times 36$. 
We randomly split the images into two sets, where each set contained five images. 
The average of peak signal-to-noise ratios (PSNR) and the average of structural similarity index measure (SSIM) of five test images were used for evaluation.
	
\textit{Network architecture \& hyperparameters.}
The CNN architectures were as following. 
$\text{CNN}_{\text F}$ was a lightweight CNN with four 2D-convolution layers. 
The first layer had $3$ and $32$ input and output channels, respectively. 
The next two layers had $32$ input and output channels. 
The last layer had $32$ input channels and $K$ output channels. 
We set $K=3$. 
$\text{CNN}_{\mu}$ also had four 2D-convolution layers, where the first layer had $3$ and $32$ input and output channels, respectively, and the rest had $32$ input and output channels, followed by a fully-connected (FC) layer with $32$ input units and $1$ output unit. 
A max pooling operator was used between two convolution layers. 
We used a Rectified Linear Unit (ReLU) after every convolution layer and FC layer.
In each experiment, we trained our proposed GTV model for 50 epochs using stochastic gradient descent (SGD) with batch size of $16$. 
The learning rate was set to $10^{-4}$. 
We set $\epsilon=0.3, \rho=0.01$, $B=6$ and $M=20$.
	
\begin{table}[ht]
\scriptsize
\centering
\ra {1.5}
\begin{tabular}{@{}ccccc@{}}
\toprule[1.2pt]
Method  & \# Parameters & $\begin{pmatrix}\sigma = 25\end{pmatrix}$ & $\begin{pmatrix}\sigma_{\text{train}}=25,\\\sigma_{\text{test}}=40\end{pmatrix}$ \\

\hline
BM3D & N/A & 30.19 $|$ 0.802 & N/A \\
CDnCNN & 0.56M & \textbf{30.39} $|$ \textbf{0.826} & 24.77 $|$ 0.482\\
            \hline
            DGLR (1 layer) &  0.45M & 30.24$|$ 0.809 & 27.27$|$ 0.742\\
            DGLR-M (1 layer) &  0.06M & 29.66 $|$ 0.774  &  26.75 $|$ 0.666 \\
            DGTV (1 layer) & 0.06M & 30.29 $|$ 0.818 & 27.68 $|$ 0.766 \\
            \hline
            DGLR (2 layers) &  0.9M & 30.36 $|$ 0.820 & 27.47  $|$ 0.763 \\
            DGLR-M (2 layers) &  0.12M & 30.29 $|$ 0.817 &  27.19 $|$ 0.731 \\
            DGTV (2 layers) & 0.12M & 30.35 $|$ 0.820 & \textbf{27.82} $|$ \textbf{0.772}\\
            \bottomrule[1.2pt]
        \end{tabular}
        \caption{Number of trainable parameters, average PSNR (left) and SSIM (right) in AWGN removal and statistical mismatch setting.}
        \label{tab:exp}
    \end{table}
    
\begin{figure}[t]
\centering
\includegraphics[width=.98\linewidth]{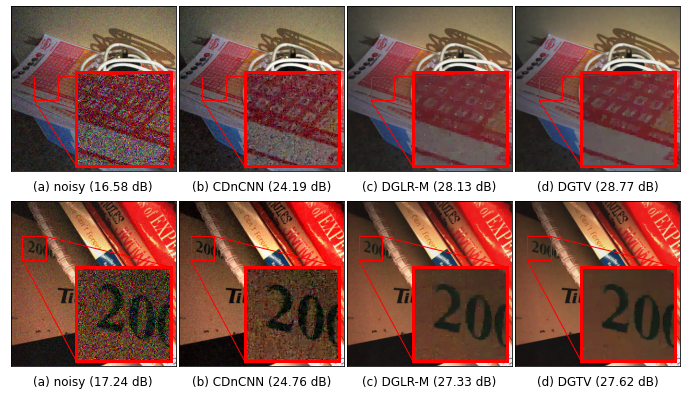}
\vspace{-0.1in}
\caption{Sample result in statistical mismatch situations. Training on $\sigma=25$ and testing on $\sigma=40$.}
\label{fig:mismatch}
\end{figure}
\subsubsection{Additive White Gaussian Noise Removal}
	
We first conducted experiments to test Gaussian noise removal ($\sigma=25$). 
To achieve comparable complexity between DGLR and DGTV, 
we replace DGLR's network architectures with the same described architectures of DGTV.
The results are shown in Table\;\ref{tab:exp}. 
Generally, learning-based methods performed better than model-based method. 
Compared to CDnCNN, DGTV achieved comparable performance (within 0.04dB in PSNR) while employing 80\% fewer network parameters. 
Compared to DGLR-M, DGTV performed better given the same number of layers.
Specifically, at 1 layer, DGTV outperformed DGLR-M by 0.63 dB in PSNR.
Fig.\;\ref{fig:gauss} shows sample results of this experiment. 
As expected, we observe that DGTV reconstructed piecewise-smooth (PWS) image patches like English letters on light background very well.

\subsubsection{Statistical Mismatch}
    
To demonstrate robustness against statistical mismatch, \ie the training and test data have different statistics, we trained all models on artificial noise $\sigma = 25$ and tested them with $\sigma = 40$. 
The final column of Table\;\ref{tab:exp} shows the results of this experiment. 
We observe that DGTV performed better than DGLR-M consistently for the same number of layers, though the gap became smaller as we increased the number of layers. 
In particular, at 1 layer, DGTV outperformed DGLR-M by 0.93dB in PSNR.
Compared to CDnCNN, our DGTV outperformed it by a large margin---more than $3$dB.
Because we employed 80\% fewer network parameters than CDnCNN, one can interpret this result on robustness against statistical mismatch to mean that our implementation is less likely to overfit training data than CDnCNN. 
Fig.\;\ref{fig:mismatch} shows sample results of this experiment. 

\section{Conclusion}
\label{sec:conclude}
Despite the recent success of deep neural network architectures for image denoising, the required large parameter set leads to a large memory footprint. 
Using a hybrid design of interpretable analytical graph filters and deep feature learning called deep GTV, we show that our denoising performance was on par with state-of-the-art DnCNN, while reducing parameter size by 80\%. 
Moreover, when the statistics of the training and test data differed, our scheme outperformed DnCNN by up to 3dB in PSNR.
As future work, we plan to investigate more frugal neural network implementations that can result in even smaller memory requirements. 




\pagebreak[4] 
\begin{small}
\bibliographystyle{IEEEbib}
\bibliography{Href}
\end{small}

\end{document}